**An Advanced Microscopic Energy Consumption Model for Automated Vehicle: Development, Calibration, Verification**

**Ke Ma**
Research Assistant
Department of Civil & Environmental Engineering
University of Wisconsin-Madison, Madison, WI, U.S., 53719
Email: kma62@wisc.edu

**Zhaohui Liang**
Research Assistant
Department of Civil & Environmental Engineering
University of Wisconsin-Madison, Madison, WI, U.S., 53719

**Hang Zhou**
Research Assistant
Department of Civil & Environmental Engineering
University of Wisconsin-Madison, Madison, WI, U.S., 53719

**Xiaopeng Li, Ph.D., Corresponding Author**
Professor
Department of Civil & Environmental Engineering
University of Wisconsin-Madison, Madison, WI, U.S., 53719
Email: xli2485@wisc.edu

Word Count: 5768 words + 5 table (250 words per table) = 7,018 words

Submission Date: August 1, 2024

*Ke Ma, Zhaohui Liang, Hang Zhou and Xiaopeng Li*

**ABSTRACT**

The automated vehicle (AV) equipped with the Adaptive Cruise Control (ACC) system is expected to reduce the fuel consumption for the intelligent transportation system. This paper presents the Advanced ACC-Micro (AA-Micro) model, a new energy consumption model based on micro trajectory data, calibrated and verified by empirical data. Utilizing a commercial AV equipped with the ACC system as the test platform, experiments were conducted at the Columbus 151 Speedway, capturing data from multiple ACC and Human-Driven (HV) test runs. The calibrated AA-Micro model integrates features from traditional energy consumption models and demonstrates superior goodness of fit, achieving an impressive 90% accuracy in predicting ACC system energy consumption without overfitting. A comprehensive statistical evaluation of the AA-Micro model's applicability and adaptability in predicting energy consumption and vehicle trajectories indicated strong model consistency and reliability for ACC vehicles, evidenced by minimal variance in RMSE values and uniform RSS distributions. Conversely, significant discrepancies were observed when applying the model to HV data, underscoring the necessity for specialized models to accurately predict energy consumption for HV and ACC systems, potentially due to their distinct energy consumption characteristics.

## INTRODUCTION

Automated vehicles (AVs) are widely regarded as one of the most promising advancements in transportation technology in the near future. Of note, researchers have highlighted the potential of AVs to plan their trajectories, thereby achieving optimal energy consumption(1–4). Trajectory tracking involves reaching and following a desired trajectory within a certain time and then maintaining motion stability continuously. Accurate and quick trajectory tracking provides practical foundations for reducing energy consumption by leveraging high-precision control systems.

Currently, most research on AV trajectory planning focuses on designing models, utilizing various trajectory planning methods (5–8). These planning models generally incorporate a term related to energy consumption, which they aim to minimize to achieve optimal trajectory. There are two prevalent options in the energy consumption term: optimizing the minimization of instantaneous acceleration squared and employing traditional energy consumption models (9–11). The first approach, which focuses on minimizing the instantaneous acceleration squared, is advantageous due to its simplicity and computational efficiency. However, its primary drawback is that it does not always guarantee the lowest energy consumption, as it oversimplifies the complex relationship between vehicle dynamics and energy consumption. On the other hand, traditional energy consumption models, such as the Virginia Tech Microscopic (VT-Micro) model, Vehicle Specific Power (VSP) model, and Australian Road Research Board (ARRB) model, although computationally more complex, offer a more accurate calculation of energy consumption (12–14). These models take into account a broader range of factors affecting energy efficiency, including vehicle speed and acceleration. By accurately modeling these factors, traditional approaches can identify more accurate energy-efficient trajectories.

However, a significant challenge arises from the fact that these traditional energy consumption models are primarily based on early data obtained from human-driven vehicles (HVs). They lack a comprehensive understanding of how energy consumption in dynamics control might change when vehicles are equipped with automated driving systems. Once the planning phase determines the optimal trajectory, the automated driving system transitions into the control phase. In the control phase, the vehicle's throttle, brake, and steering inputs are adjusted to follow the planned optimal trajectory. This process involves multiple control functions that simulate human driving behavior, e.g., proportional-integral-derivative (PID) function or model prediction control (MPC) function (15–17). Nevertheless, it remains uncertain whether these control functions can be considered equivalent to human driving actions.

The discrepancy between human and automated driving systems introduces uncertainties in the accuracy of existing energy consumption models when applied to AVs. While human drivers might exhibit variability in their control inputs due to factors such as reaction time and decision-making processes, automated driving systems are designed to perform with high precision and consistency. This is particularly evident when adapting the MPC function in dynamics to mitigate control oscillations. This fundamental difference could lead to variations in energy consumption patterns that traditional models do not account for. Therefore, it is crucial to develop new energy consumption models tailored specifically for AV to ensure that the predicted energy savings are realized in practical applications.

Among the various systems enhancing AV functionality, the Adaptive Cruise Control (ACC) system is considered to have a significant impact on longitudinal driving behavior, which is widely integrated into commercial AVs. This system autonomously adjusts the vehicle's speed to maintain a safe following distance from the preceding vehicle. By leveraging fast-responding onboard computers and sensors, the ACC system achieves high precision and stability in speed control, outperforming human drivers who exhibit uncertain and unpredictable behaviors. Thus, some researchers believe that ACC-equipped vehicles (hereinafter referred to as ACC vehicles) have the potential to reduce overall energy consumption with widely used smoothed MPC functions in commercial ACC vehicles.

However, this perspective is not without contention. Some researchers argue that the automation inherent in ACC systems does not come without costs. The reliance on onboard sensing and computing devices may lead to additional energy consumption due to increased weight, computing load, sensor load, and aerodynamic drag. Several studies have been conducted to investigate the energy impacts of ACC vehicles using simulations. While simulations under ideal conditions—often requiring strong

assumptions—tend to show positive outcomes regarding energy efficiency improvement, some tests about sensing and computing devices subjected to real-world constraints yield less clear results. These debates highlight the need for comprehensive field experiments to test and verify the actual energy benefits of ACC vehicles under practical conditions, especially compared with HVs'.

Current studies have included limited field experiments to investigate the commercial ACC vehicle's performance. A recent study, including 14 ACC vehicle datasets, detailed introduced field experiments and data collection process. For instance, Gunter et al. (18) studied the string stability of ACC vehicles by calibrating an optimal velocity model with a set of field data. Shi and Li (19) developed traffic flow fundamental diagrams for AVs under different penetration rates based on field experiment data. Makridis et al. conducted a field experiment with five commercial ACC vehicles to examine their impacts on traffic flow and string stability (20). However, these studies predominantly focused on ACC vehicle car-following behavior modeling, string stability analyses, and traffic system impacts, while the aspect of ACC vehicle energy consumption is more limited due to demanding experimental conditions. Some studies conducted by Knoop (21) are relatively relevant to this paper. They performed a field experiment with seven commercial ACC vehicles driving as a platoon over a distance of nearly 500 km. Utilizing a classical energy consumption model, they categorized vehicles based on their total emission contribution. However, they did not collect actual energy consumption data, instead calculating energy consumption based on traditional HV energy models. While this analysis provides some insights into ACC vehicle energy consumption, it is limited by the lack of actual energy consumption data.

In contrast, our study aims to address this gap by conducting experiments that collect actual trajectory and energy consumption data, thereby providing a more accurate assessment of energy consumption in ACC vehicles. This study first collected trajectory data of commercial ACC vehicles including speed and acceleration information. Specifically, we assessed energy consumption by measuring combustion engine and electric motor data through the On-Board Diagnostics (OBD) system and proposed a general energy consumption function. Utilizing this data, we proposed an advanced VT-Micro model tailored for the ACC vehicle, achieving a high degree of fitness. Additionally, we collected data for HV under the same conditions and using the same vehicle platform, allowing for a direct comparison between the two energy consumption modes. This comprehensive approach enables a more accurate evaluation of energy consumption in the ACC system and provides a foundation for more consumption studies for AVs.

## DATA COLLECTION

This section introduces the test vehicle platform, experiment setting, and data collection.

### Test Platform

The ACC vehicle platform is developed by the CATS Lab. This ACC vehicle platform, based on a 2016 Lincoln MKZ hybrid, is designed to facilitate Level 3 automation, which allows for full control of the vehicle under certain conditions while still requiring human override capability. This ACC vehicle platform features a 2.0-liter Atkinson-cycle inline-4 hybrid engine including Internal Combustion Engine (ICE) and Electric Engine (EE) power, which produce 188 horsepower. It combines a gasoline engine and an electric motor. Note that the ACC vehicle can also be operated by human drivers by disabling the ACC systems, i.e., HVs in the following. The ACC vehicle platform activates at a speed of 20 miles per hour (mph), and all data collection occurs only at speeds above this threshold. The speed upper limit is 40 mph. Each test run involved the same driver in the vehicle (including ACC tests), with the vehicle's weight distribution kept constant.

### Experiment Setting

The experiment was conducted at the Columbus 151 Speedway test site located in Columbus, Wisconsin. The site features a 500 meters almost straight road, ideal for consistent testing conditions. Our experiment included 9 tests with the ACC system control and 9 tests with the HV control. All tests were conducted with the vehicle's lights and air conditioning set to the same levels. Each test included a start-stop at a fixed location to maintain uniformity in the experimental conditions. In the ACC vehicle tests, each run has the same headway and speed settings. In the HV tests, the driver tried to keep the driving speed as similar as possible in each test.

**Data Collection**

We introduce a method utilizing Original Equipment Manufacturer (OEM) sensors to gather energy consumption data from both the ICE and EE, standardizing this data into joules for comparative analysis. Fuel usage is assessed by measuring the equivalent air intake of the engine, while the energy consumption of the electric motor is determined by monitoring the state of charge (SOC) of the high-voltage battery. Data access is facilitated through the OBD port, allowing non-intrusive data collection while maintaining the integrity of the vehicle's systems. The adoption of the OBD protocol, OBD-II evolved in 1994 and became compulsory for all passenger vehicles in the United States by 1996. Currently, as stipulated in ISO 15765, OBD scanners are integrated within the vehicle's Controller Area Network (CAN) BUS system. A standard CAN message includes several functional blocks to ensure data accuracy, with the primary elements being the message identifier and the data block. In typical diagnostics, a query with a specific CAN ID and data format, defined by the Society of Automotive Engineers (SAE) standard J1979, is sent. The vehicle's Electronic Control Unit then retrieves and returns the requested data from the designated CAN ID plus 8. By retrieving this data, the instant speed, acceleration, and energy consumption data can be obtained by the OBD scanner.

The ACC vehicle platform, experiment site, and OBD scanner are shown in Figure 1.

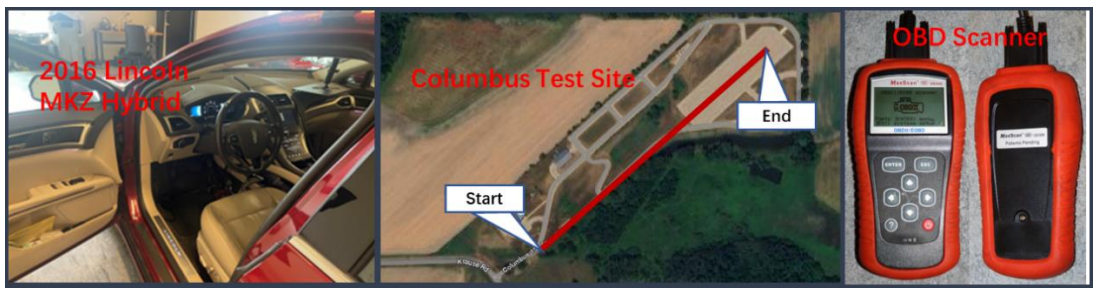

**Figure 1 ACC vehicle platform by 2016 Lincoln MKZ hybrid, experiment site in Columbus, and OBD scanner.**

## PROBLEM FORMULATION AND DATA PROCESS

**Problem Formulation**

Before delving into the details of the energy and trajectory analysis, it is important to formulate this problem and define notations.

Define $i^{\{a,h\}} \in \mathcal{J}^{\{a,h\}} := \{1,2,\cdots,I^{\{a,h\}}\}$ to represent the ACC vehicle and HV trajectory index set, where $I^a = 9$ and $I^h = 9$ refers to the total number of ACC vehicle and HV trajectories in this paper. Each trajectory $i$ contains actual data recorded at consecutive time points $t \in \mathcal{T} := \{0,1,\cdots,T\}$, and $T \in \mathbb{R}$ is the total number of time indexes. Each two consecutive time points is separated by a constant unit time interval, denoted by $\Delta t = 1$ seconds in this paper.

**Data Process**

To analyze the energy data, we proposed a data process of energy consumption and a summary of the final cleaned data statistics.

The originally collected data from the OBD scanner involves monitoring the following signals: Mass Air Flow (MAF) (unit: g/s), Battery State of Charge (SOC) (unit: %), and Vehicle Speed (unit: m/s), denoted as $MAF$, $SOC$, and $v_t$, respectively. Although these data are collected adhering to standard protocols (SAE J1979 and J2190), with a collection frequency of approximately 20 milliseconds, the response time of many actuators/sensors cannot match this frequency. Thus, the data are integrated and recorded at $\Delta t = 1$ s intervals to ensure accuracy and consistency.

The first step is to process the energy consumption data. The energy consumption data is processed using both engine and electric motor data, considering the hybrid nature of the vehicle. Thus, all energy model calculations are ultimately converted to J/s. The computation is divided into two main components: the energy consumption of gasoline from ICE and electric power in EE, denoted as $J_t^{\mathrm{g}}$ and $J_t^{\mathrm{e}}$, respectively. The calculation of $J_t^{\mathrm{g}}$ is shown in Equation 1,

$$J_t^{\mathrm{g}} = \int_t^{t+\Delta t} \frac{MAF \times \lambda^g \times \eta^{\mathrm{g}}}{\lambda^{\mathrm{a}}} \, , t \in \mathcal{T} \tag{1}$$

where $\lambda^g, \lambda^{\mathrm{a}}$ and $\eta^{\mathrm{g}}$ refer to the air-fuel ratio, calorific value of gasoline, and gasoline in engine operation efficiency, respectively. *MAF* is recorded from the OBD scanner. Define the energy consumption of gasoline set as $\mathcal{J}^{\{\mathrm{a,h}\},\mathrm{g}} := \left\{ J_t^{\{\mathrm{a,h}\},\mathrm{g}} | t \in \mathcal{T} \right\}$.

The calculation of $m_b$ is shown in Equation 2,

$$J_t^{\mathrm{e}} = \int_t^{t+\Delta t} (SOC_t - SOC_{t+\Delta t}) C_0 \times \eta^{\mathrm{e}} \, , t \in \mathcal{T} \tag{2}$$

where $C_0$ and $\eta^{\mathrm{e}}$ are battery capacity and electric power operation efficiency, respectively. Define the energy consumption of electric power set as $\mathcal{J}^{\{\mathrm{a,h}\},\mathrm{e}} := \left\{ J_t^{\{\mathrm{a,h}\},\mathrm{e}} | t \in \mathcal{T} \right\}$. The values of parameters in Equation 1 and 2 can be found in Table 1.

**TABLE 1 Parameter values in $J_t^{\mathrm{g}}$ and $J_t^{\mathrm{e}}$ calculation**

| Parameter | $\lambda^{\mathrm{a}}$ | $\lambda^g$(J/g) | $C_0$ (kWh) | $\eta^{\mathrm{g}}$ | $\eta^{\mathrm{e}}$ |
|---|---|---|---|---|---|
| Value | 14.7 | 44,000 | 1.4 | 40% | 50% |

(The battery capacity of 1.4 kWh is equal to 5,040,000 J).

Thus, the total energy consumption ($J^{\mathrm{t}}$), i.e., the sum of the energy consumption of $J_t$ and $J^{\mathrm{b}}$, is shown in Equation 3. Similarly, we define the total energy consumption set as $\mathcal{J}^{\{\mathrm{a,h}\}} := \left\{ J_t^{\{\mathrm{a,h}\}} | t \in \mathcal{T} \right\}$.

$$J_t = J_t^{\mathrm{g}} + J_t^{\mathrm{e}} \tag{3}$$

The second step is to process the trajectory data. We need to exclude data points where the speed is less than 20 miles/hour (approximately 8.941 m/s) because the ACC system is only activated at speeds higher than this value in this test platform.

Then, the accelerations $a_t$ are calculated by the difference of velocities as shown in Equation 4,

$$a_t = \frac{v_t - v_{t-1}}{\Delta t}, t \in \mathcal{T}/0 \tag{4}$$

Define the velocity and acceleration set as $\mathcal{V}^{\{\mathrm{a,h}\}} := \left\{ v_t^{\{\mathrm{a,h}\}} | t \in \mathcal{T} \right\}$ and $\mathcal{A}^{\{\mathrm{a,h}\}} := \left\{ a_t^{\{\mathrm{a,h}\}} | t \in \mathcal{T} \right\}$, respectively.

Also, the data points need to be further cleaned since the differential process introduces some significant errors in acceleration. We need to remove the data index where the acceleration $a_t$ is zero and the total energy $J_t$ is less than 25,000 J. The rationale behind this criterion stems from the characteristics of the ACC vehicle platform being studied. When the ACC vehicle's acceleration is zero, it indicates that the ACC vehicle is in a steady state, either stationary or moving at a constant speed. During these conditions, the vehicle's propulsion relies solely on the electric motor drive, bypassing the typical hybrid energy formula that integrates both the internal combustion engine and the electric motor. The threshold of 25,000 J is derived from empirical observations. It considers the maximum battery capacity and the energy required to maintain the vehicle's balance during steady states. This anomaly in energy consumption necessitates the exclusion of such data points to maintain the validity of our analysis.

**Data Statistics**

    After cleaning, the dataset includes 2402 data points (about 40 minutes of data) for the ACC vehicle and 2021 data points (about 34 minutes of data) for the HV. Table 2 summarizes the cleaned data variable after the data processing and basic statistics, including the minimum, maximum, mean, and standard deviation (Std) values of the cleaned data.

**TABLE 2 Cleaned data variable and basic statistics.**

| Data Variable | Description | Unit | Range [min, max] | Mean | Std |
|---|---|---|---|---|---|
| Trajectory Index | Trajectory Index for ACC vehicle and HV $\mathcal{J}^{\{a,h\}}$ | N/A | $\mathcal{J}^a = [1,9]$ $\mathcal{J}^h = [1,9]$ | N/A | N/A |
| Time | Time index with fixed sampling intervals. | s | N/A | N/A | N/A |
| Battery_J | Energy consumption of electric power in EE | J | $\mathcal{J}^{a,e} = [-15027.8, 11769.8]$ $\mathcal{J}^{h,e} = [-15907.1, 8064.0]$ | 2711.3 2574.9 | 3204.5 3291.3 |
| Engine_J | Energy consumption of gasoline from ICE | J | $\mathcal{J}^{a,g} = [1041.6, 50767.0]$ $\mathcal{J}^{h,g} = [1041.6, 572870.0]$ | 27831.0 24583.3 | 15580.1 18462.8 |
| Total_J | Total energy consumption | J | $\mathcal{J}^a = [-13979.4, 50767.0]$ $\mathcal{J}^h = [-14666.2, 56845.6]$ | 25119.6 22008.5 | 17828.2 20645.6 |
| Speed | Speed of the following vehicle. | m/s | $\mathcal{V}^a = [8.957, 15.278]$ $\mathcal{V}^h = [8.941, 17.778]$ | 12.996 12.637 | 1.867 2.091 |
| Acceleration | Acceleration of the following vehicle. | m/s$^2$ | $\mathcal{A}^a = [-5.390, 3.057]$ $\mathcal{A}^h = [-4.632, 2.897]$ | -0.027 0.000 | 1.532 1.619 |

    It is noted that the value of 'Battery_J' can be negative due to the processes of charging and discharging, indicating that the ACC vehicle may be engaging in regenerative braking during this time. Thus, it leads that the value of 'Total_J' can also be negative.

**ENERGY CONSUMPTION MODEL**

    In this section, two traditional vehicle fuel consumption models are introduced first, including the VT-Micro model and the ARRB model. However, these models were developed many years ago based on older vehicle chassis tests. To address the limitations of these traditional models, we synthesized their characteristics and proposed an advanced energy consumption model, the Advanced ACC Micro (AA-Micro) model. The AA-Micro model integrates features from both the VT-Micro and ARRB models, demonstrating higher fitness and improved accuracy in predicting energy consumption for the tested vehicle, especially when it is under ACC system control.

**Traditional Energy Consumption Model**

    The VT-Micro model, developed by Ahn et al. (22), is designed to measure the instantaneous fuel consumption of individual vehicles. The model utilizes instantaneous speed and acceleration as inputs to

predict fuel consumption. The model shown in Equation 5 utilizes instantaneous speed and acceleration as inputs to predict fuel consumption.

$$F_t^{\text{VTM}} = \exp\left(\sum_{n_1=0}^{3}\sum_{n_2=0}^{3} f_{n_1 n_2}^{\text{VTM}}(v_t)^{n_1}(a_t)^{n_2}\right) \tag{5}$$

where $n_1, n_2$ are the power indexes and $f_{n_1 n_2}^{\text{VTM}}$ are constant coefficients for each term in the VT-Micro model.

The ARRB model, developed by Akcelik (23), is an elemental model based on vehicle operation modes such as cruising, deceleration, idling, and acceleration shown in Equation 6.

$$F_t^{\text{ARRB}} = f_1^{\text{ARRB}} + f_2^{\text{ARRB}}v_t + f_3^{\text{ARRB}}v_t^2 + f_4^{\text{ARRB}}v_t^3 + f_5^{\text{ARRB}}v_{it}\cdot a_t + f_6^{\text{ARRB}}v_t(\max(0, a_t)^2) \tag{6}$$

where $f_{\{1,\dots,6\}}^{\text{ARRB}}$ are corresponding parameters in the ARRB model.

It is worth noting that the coefficients may have significant differences from the original literature. The reason is that we set up the output as J which is different from L/mL in literature. This does not affect the performance of the final calibration.

### AA-Micro Model

The VT-Micro model has a rich combination of terms, allowing it to capture various factors influencing energy consumption. Its nonlinearity, driven by exponential terms, represents the complexity of human control. In contrast, the ARRB model highlights the asymmetry and linearity in energy consumption due to positive acceleration and linear structure, a factor neglected by the VT-Micro model. However, the ARRB model overlooks other potential influences. To address these limitations, we propose the AA-Micro model shown in Equation 7, which integrates the strengths of both models in asymmetry and multiple terms:

$$\begin{aligned}F_t^{\text{AAM}} = \sum_{n_1=0}^{2}\sum_{n_2=0}^{2} \big(&f_{n_1 n_2}^{\text{AAM}}(v_t)^{n_1}(a_t)^{n_2} + f_{n_1 n_2}^{\text{p}}(\max(0, v_t))^{n_3}(\max(0, a_t))^{n_4} \\ &+ \exp\big(g_{n_1 n_2}^{\text{AAM}}(v_t)^{n_1}(a_t)^{n_2} + g_{n_1 n_2}^{\text{p}}(\max(0, v_t))^{n_3}(\max(0, a_t))^{n_4}\big)\big)\end{aligned} \tag{7}$$

where $n_1$ and $n_2$ are the power indexes in the AA-Micro model. $f^{\text{AAM}}, f^{\text{p}}, g^{\text{AAM}},$ and $g^{\text{p}}$ are corresponding linear/non-linear parameters for all data and positive in the AA-Micro model, respectively. This model captures a broader range of factors, including the asymmetry in acceleration effects of ACC control. Also, the AA-Micro model employs a combination of linear and nonlinear structures. This showed a noticeable improvement in fitness during testing. We reduced the order of $n_1$ and $n_2$ to the second degree compared with the VT-Micro model, as we found that third-degree terms caused significant overfitting. Furthermore, we included both linear and nonlinear terms because each term showed a noticeable improvement in fitness during testing.

### Calibration and Verification

This section introduces how we calibrate different energy consumption models in the training data set $\mathcal{J}^{\text{Train},\{a,h\}}$ and verify them in the test data set $\mathcal{J}^{\text{Test},\{a,h\}}$. The training data set $\mathcal{J}^{\text{Train},\{a,h\}}$ consists of approximately 80 percent of the cleaned data points in $\mathcal{J}^{a,h}$, calculated as $\left|\mathcal{J}^{\text{Train},\{a,h\}}\right| = \left\lfloor 0.8 \times \left|\mathcal{J}^{\{a,h\}}\right|\right\rfloor$, where $\lfloor\cdot\rfloor$ denotes the floor function. The test data set $\mathcal{J}^{\text{Test},\{a,h\}}$ consists of the remaining data points, i.e., $\mathcal{J}^{\text{Test},\{a,h\}} = \mathcal{J}^{\{a,h\}}/\mathcal{J}^{\text{Train},\{a,h\}}$. Let $\theta^{\{\text{VTM,ARRB,AAM}\},\{a,h\}}$ denotes the parameters for each model.

We fitted these models using linear regression by the least squares error method, which aims to minimize the sum of squared errors for all data points in $\mathcal{J}^{\text{Test},\{a,h\}}$. Let $r_t^{\{a,h\}} := \hat{J}_t^{\{a,h\}} - J_t^{\{a,h\}}$ denote the residual, reflecting the difference between the actual and predicted energy consumption of each model. Based on the definition, the optimization for $\theta_*^{\{\text{VTM,ARRB,AAM}\},\{a,h\}}$ is defined as:

$$\theta_*^{\{\text{VTM,ARRB,AAM}\},\{a,h\}} := \mathrm{argmin}_{\theta^{\{\text{VTM,ARRB,AAM}\},\{a,h\}}} \frac{1}{\left|\mathcal{J}^{\text{Train},\{a,h\}}\right|} \sum_{i \in \mathcal{J}^{\text{Train},\{a,h\}}} \left(r_t^{\{a,h\}}\right)^2 \quad (8)$$

After the model calibration, the test data were employed to validate the proposed MPL model, assess its reliability, and check for overfitting, which is a key concern with regression-based models. To evaluate the risk of overfitting, we compared adjusted R-squared ($R_{\text{adj}}^2$) values for the calibration data, verification data, and the entire updated dataset.

The results of the calibration and verification for each model are presented in Table 3. The $R_{\text{adj}}^2$ of all cleaned data is also recorded in Table 3.

**TABLE 3 $R_{adj}^2$ Comparison of traditional energy consumption model and AA-Micro Model.**

| Statistical measure | Calibration | Verification | Total Cleaned Data |
|---|---|---|---|
| $R_{\text{adj}}^2$ of VT-Micro model | $\mathcal{J}^{\text{Train,a}}$: 0.620 | $\mathcal{J}^{\text{Test,a}}$: 0.568 | $\mathcal{J}^{\text{a}}$: 0.772 |
| | $\mathcal{J}^{\text{Train,h}}$: 0.662 | $\mathcal{J}^{\text{Test,h}}$: 0.582 | $\mathcal{J}^{\text{h}}$: 0.808 |
| $R_{\text{adj}}^2$ of ARRB model | $\mathcal{J}^{\text{Train,a}}$: 0.736 | $\mathcal{J}^{\text{Test,a}}$: 0.712 | $\mathcal{J}^{\text{a}}$: 0.792 |
| | $\mathcal{J}^{\text{Train,h}}$: 0.689 | $\mathcal{J}^{\text{Test,h}}$: 0.597 | $\mathcal{J}^{\text{h}}$: 0.736 |
| $R_{\text{adj}}^2$ of AA-Micro model | $\mathcal{J}^{\text{Train,a}}$: 0.895 | $\mathcal{J}^{\text{Test,a}}$: 0.836 | $\mathcal{J}^{\text{a}}$: 0.902 |
| | $\mathcal{J}^{\text{Train,h}}$: 0.890 | $\mathcal{J}^{\text{Test,h}}$: 0.802 | $\mathcal{J}^{\text{h}}$: 0.887 |

We can notice that the AA-Micro model consistently provides the best estimation of energy consumption across all conditions — whether for HVs or ACC vehicles, and in both training and testing datasets. Notably, the AA-Micro model shows no signs of overfitting, as indicated by its high $R_{\text{adj}}^2$ values.

The VT-Micro model shows relatively high fitness for HV predictions but performs less well for ACC vehicles. This is likely due to its nonlinear structure better capturing HV's stochastic behavior. The ARRB model performs well in predicting AV energy consumption, benefiting from its linear structure. However, it is less effective than the AA-Micro model in overall performance and more prone to underfitting in HVs. The AA-Micro model demonstrates superior performance in both HV and ACC predictions. Its combined linear and nonlinear structure accounts for the complexities of both vehicle types, leading to the highest $R_{\text{adj}}^2$ across all datasets.

Figure 2 illustrates the actual energy consumption compared to predictions from the AA-Micro, ARRB, and VT-Micro models for ACC vehicles in (a) and HV in (b). The blue line represents actual energy consumption, while the predictions from the AA-Micro, VT-Micro, and ARRB models are shown with dashed green, dotted yellow, and dashed red lines, respectively. From the figure, it is evident that the AA-Micro model provides the closest predictions to the actual energy consumption in both ACC and HV cases, demonstrating its superior performance and minimal overfitting. The VT-Micro model, while capturing the stochastic behavior of HVs, does not perform as well with ACC vehicles. The ARRB model shows reasonable performance, particularly in predicting AV energy consumption, likely due to its linear structure.

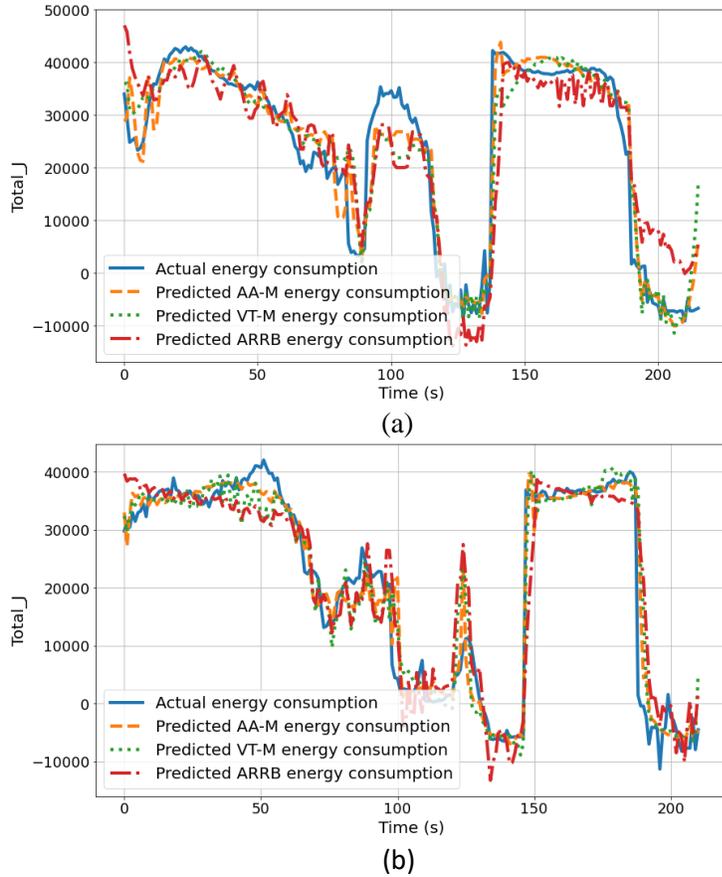

(a)

(b)

**Figure 2 Actual energy consumption compares with AA-Micro, ARRB, and AA-Micro model predictions for ACC vehicle in (a) and HV in (b). (Trajectory $i^a = 3$ and $i^h = 3$).**

## MODEL ADAPTABILITY TEST

To ensure the actual utility of the AA-Micro model, it must be tested for adaptability to ensure they are effective and reliable in reflecting the behaviors of ACC vehicles in different test runs. The adaptability tests in this section focus on three tests:

- Test 1: AA-Model consistency in ACC vehicle. To ensure the practical utility of the AA-Micro model, we assess its adaptability by fitting it to the trajectory data from one ACC test run and then testing it on other ACC vehicle test runs. Statistical significance in these applications can indicate the model's reliability in different scenarios.

- Test 2: AA-Model distinction in ACC vehicle and HV. To demonstrate that HV and ACC vehicles cannot share the same underlying model. This distinction underscores the AA-Micro model's specificity to ACC vehicle behavior.

Due to the large amount of data, we simplified the work by combining into 3 data groups from 9 ACC vehicle and HV test runs. Define group index $j = \{1,2,3\}$. The group $j = \{1,2,3\}$ include [1,4,7], [2,5,8], and [3,6,9] test runs, respectively.

To assess the adaptability of the AA-Micro model, we employed Root Sum Square (RSS) and Root Mean Squared Error (RMSE) metrics. These statistical measures determine whether variations in the models' outcomes for different ACC vehicles are different. In this assessment, the RSS and RMSE values for the total energy consumption predicted by the proposed AA-Micro model are used, as shown in Equations 8 and 9.

$$RSS_j^{\{a,h\}} = \sum_{t_j}^{T_j} \left( r_{t_j}^{\{a,h\}} \right) \tag{9}$$

$$RMSE_j^{\{a,h\}} = \frac{1}{T_i} \sum_{t_j=1}^{T_j} \sqrt{\left( r_{t_j}^{\{a,h\}} \right)^2} \tag{10}$$

The results of these metrics are computed for the AA-Micro model when it is fitted to the trajectory data from one ACC test run and tested on other ACC vehicle test runs. A comparison of these values provides an indication of the model's reliability and robustness in different scenarios. By comparing the MSE and MAE values across different groups, we can demonstrate whether the AA-Micro model contains a high degree of consistency and reliability.

**Test 1: AA-Model Consistency in ACC Vehicle**

First, Figure 3 displays the distribution of RSS for three ACC vehicle group's AA-Micro models applied to each ACC vehicle dataset. RSS denotes the differences between observed ACC vehicle energy consumptions and the values predicted by the AA-Micro models calibrated by ACC vehicle data. These figures illustrate how well the three AA-Micro models predicted the energy consumption of the tested ACC vehicle. Each line on the graph represents the density of residuals for the corresponding model, with the ideal model having RSS clustered closely around zero and with minimal spread, indicating predicted values close to the actual values. The Std shows how spread out the residuals were. A smaller Std. indicates the residuals were closer to the mean, typically reflecting a better model fit.

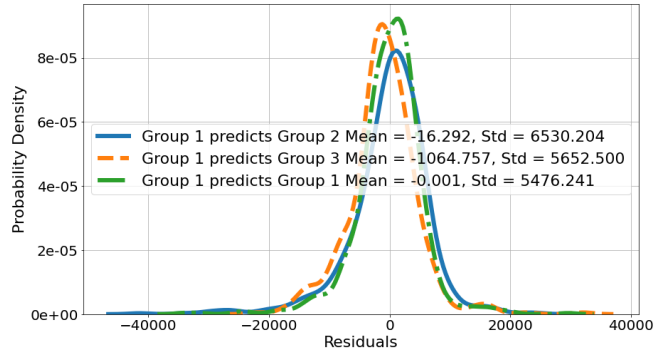

(a)

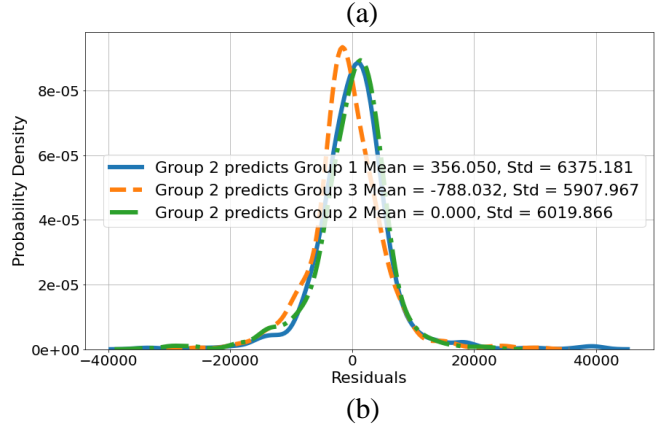

(b)

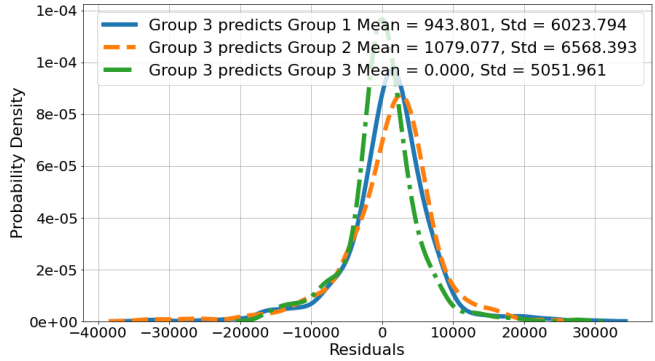

(c)

**Figure 3 RSS distribution of each AA-Micro model applied to each ACC group data. (a). group 1 ACC model prediction in each ACC group data. (b). group 2 ACC model prediction in each ACC group data. (c). group 3 ACC model prediction in each ACC group data.**

**TABLE 4 *RMSE* comparison of each AA-Micro model applied to each ACC group data.**

|                | Group 1 data | Group 2 data | Group 3 data |
|----------------|--------------|--------------|--------------|
| **Group 1 model** | 6526.147 | 5745.895 | 5472.864 |
| **Group 2 model** | 6381.191 | 5953.950 | 6016.107 |
| **Group 3 model** | 6093.613 | 6652.393 | 5046.491 |

From the probability density plots in Figure 3, the similarity in the residual distributions suggests that the model does not show significant bias when predicting different data sets. Furthermore, the RMSE values in Table 4 highlight that the differences in prediction errors among the models are minimal. Each model's RMSE values for predicting different groups are close, reinforcing the uniformity and reliability of the AA-Micro model across varied test runs. Thus, the analysis confirms that the AA-Micro model demonstrates strong adaptability and consistency, making it a robust tool for predicting ACC vehicle trajectories in diverse conditions.

**Test 2: AA-Model Distinction in ACC Vehicle and HV.**

Figure 4 displays the distribution of RSS for three ACC vehicle group's AA-Micro models applied to each HV dataset. Thus, this RSS denotes the differences between observed HV energy consumptions and the values predicted by the AA-Micro models calibrated by ACC vehicle data. This result is intended to show the statistical difference between the HV data and the ACC model predictions.

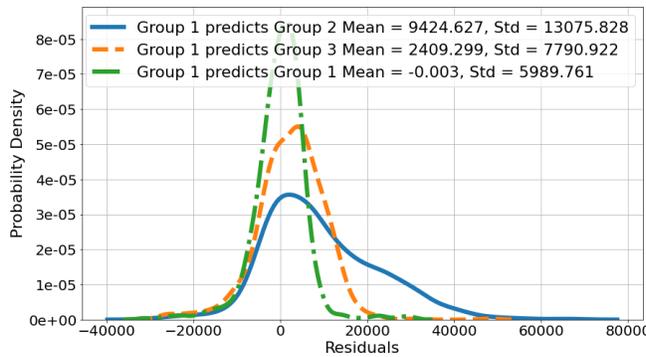

(a)

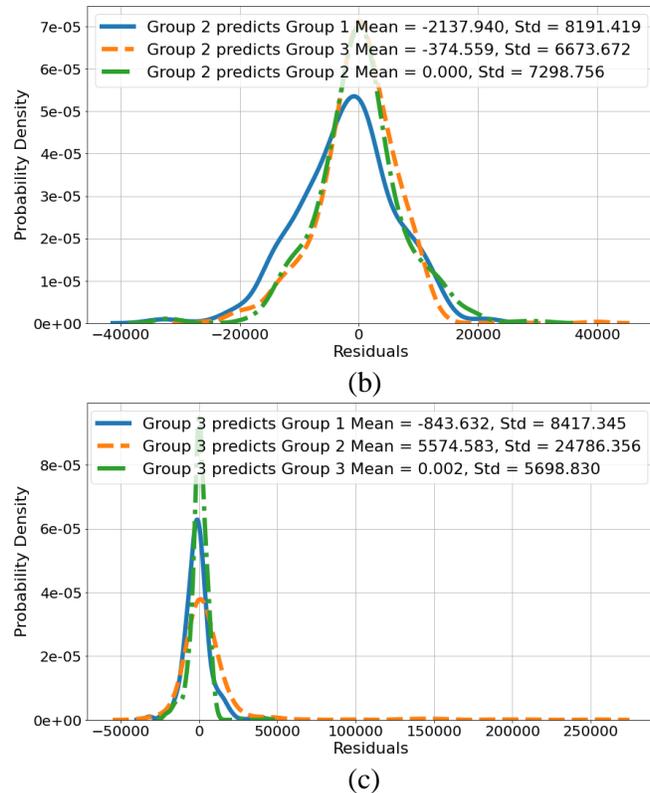

(b)

(c)

**Figure 4 RSS distribution of each AA-Micro model applied to each HV group data. (a). group 1 ACC model prediction in each HV group data. (b). group 2 ACC model prediction in each HV group data. (c). group 3 ACC model prediction in each HV group data.**

**TABLE 5 *RMSE* comparison of each AA-Micro model applied to each HV group data.**

|               | Group 1 data | Group 2 data | Group 3 data |
|---------------|-------------|-------------|-------------|
| **Group 1 model** | 16108.234   | 8459.417    | 8452.748    |
| **Group 2 model** | 8149.425    | 6679.230    | 25382.458   |
| **Group 3 model** | 5984.921    | 7291.801    | 5694.600    |

From the probability density plots in Figure 4, we observe that the means and standard deviations of the residuals for each group prediction are substantially different from zero and exhibit more spread out and diverse distribution shapes. This significant deviation in the residual distributions suggests that the model is biased and less accurate when predicting HV data. Furthermore, the RMSE values in Table 5 highlight that the differences in prediction errors among the models are substantial. Each model's RMSE values for predicting HV data are significantly higher compared to their RMSE values for predicting ACC data. This substantial increase in RMSE values reinforces the disparity and indicates that the AA-Micro model is not well-suited for predicting HV energy consumption.

## CONCLUSIONS

The ACC vehicle platform used for this study was developed by the CATS Lab, based on a 2016 Lincoln MKZ hybrid designed to facilitate Level 3 automation. This platform, featuring a hybrid engine and the capability for both ACC and human-driven control, was tested at the Columbus 151 Speedway test site in Columbus, Wisconsin. Energy consumption data were collected using OEM sensors and standardized to joules for comparative analysis. Data collection was facilitated through the OBD port, capturing metrics such as MAF, SOC, and vehicle speed. The collected data were integrated at one-second intervals to ensure accuracy and consistency. Energy consumption was calculated separately for the ICE and EE, and then summed for total energy consumption.

The AA-Micro model was developed to address the limitations of traditional models like VT-Micro and ARRB, which were based on older vehicle chassis tests. By integrating features from both models, the AA-Micro model aims to provide higher accuracy and better fit for energy consumption predictions, particularly under ACC system control. The model incorporates both linear and nonlinear terms to capture a broad range of influencing factors, including the asymmetry in acceleration effects of ACC control. Model calibration was performed using linear regression by the least squares error method, minimizing the sum of squared errors. The calibration and verification process involved splitting the data into training and test sets and evaluating the models' performance using $R^2_{\text{adj}}$.

In statistics tests, the AA-Micro model demonstrates strong adaptability and consistency when applied to ACC vehicle data, making it a robust tool for predicting ACC vehicle trajectories. However, its application to HV data underscores the necessity for customized models tailored to the unique characteristics of different vehicle types. The significant discrepancies in the model's performance for HV predictions indicate that HVs require distinct modeling approaches to ensure accurate and reliable predictions. This study emphasizes the importance of context-specific model training and validation to enhance the reliability of autonomous and semi-autonomous vehicle systems across diverse operational environments.

## AUTHOR CONTRIBUTIONS

The authors confirm their contribution to the paper as follows: study conception and design: Ke Ma; data collection: Zhaohui Liang; data analysis: Hang Zhou; All authors reviewed the results and approved the final version of the manuscript.